\documentclass[conference]{IEEEtran}
\usepackage{cite}
\usepackage{amsmath,amssymb,amsfonts}
\usepackage{algorithmic}
\usepackage{graphicx}
\usepackage{textcomp}
\usepackage{xcolor}
\usepackage[hyphens]{url}
\usepackage[hidelinks]{hyperref}
\usepackage{microtype}
\usepackage{stfloats}
\usepackage{booktabs}
\def\BibTeX{{\rm B\kern-.05em{\sc i\kern-.025em b}\kern-.08em
    T\kern-.1667em\lower.7ex\hbox{E}\kern-.125emX}}
\begin{document}

\title{Multi-objective drift-aware MLOps agent for model retraining\\
\thanks{California State University, Fullerton}
}

\author{\IEEEauthorblockN{Kumuda Valli V}
\IEEEauthorblockA{\textit{Computer Science} \\
\textit{California State University, Fullerton}\\
Fullerton, United States of America\\
vk.valli2909@gmail.com}
\and
\IEEEauthorblockN{Anli Ji}
\IEEEauthorblockA{\textit{Computer Science} \\
\textit{California State University, Fullerton}\\
Fullerton, United States of America\\
anli.ji@fullerton.edu}
}
\maketitle

\begin{abstract}
Deployed ML models face non-stationary data and delayed labels, yet most retraining strategies rely on fixed schedules or brittle drift thresholds that ignore operational cost. In this setting, retraining is a sequential decision problem rather than a one-time threshold choice. We present \textit{DARE} (Drift-Aware Retraining Engine), an offline, event-driven retraining framework that separates ingestion, monitoring, policy decisions, training, and logging through explicit schemas and deterministic replay. 

We compare DARE against practical heuristic retraining baselines, including fixed-interval, threshold-persistence, and action-rate-matched fixed scheduling policies. On Airline, the locked LinUCB policy achieves a higher mean F1 while substantially reducing intervention frequency relative to heuristic and action-rate matched baselines. Additional runtime-calibrated evaluation shows that these gains persist when measured using the measured cumulative update wall-clock time. These results suggest that conservative sequential retraining policies can improve performance while reducing unnecessary operational overhead under delayed feedback.
\end{abstract}


\section{Introduction}
Deployed machine learning models degrade when data distributions shift over time, and retraining too frequently is expensive in compute, engineering effort, and operational risk \cite{b1,b2}. In production, teams often use fixed schedules or static drift thresholds to trigger updates. These heuristics are simple but brittle: a setting that works in one regime can over-trigger or under-trigger in another \cite{b3,b4}.

This work treats retraining as a sequential decision problem with explicit operational constraints. Instead of asking whether drift crossed a threshold at one instant, we ask which action should be taken now (\textit{defer} or \textit{fine-tune}) given current drift/performance signals and expected operational overhead, while accounting for delayed feedback \cite{b5,b6}. This framing is motivated by recent ML systems work showing that retraining policy quality depends on balancing utility and pipeline operational overhead, not detection alone \cite{b7,b8},\cite{b9}.

 In this paper, \emph{MLOps} (Machine Learning Operations) refers to the engineering practice of deploying, monitoring, and maintaining machine learning systems in production. We use \emph{events} to denote timestamped records emitted by pipeline stages (e.g., monitoring outputs and execution outcomes). A \emph{policy} is a decision rule that maps the current system state to an \emph{action}, where actions in our setting include \texttt{defer} and \texttt{fine\_tune}.

We implement this framing in \textit{DARE}, an offline event-driven architecture with deterministic replay and schema-validated event contracts. The architecture decouples monitoring from decision and execution, enabling reproducible, policy-level comparisons under matched data windows and delay conditions.

We evaluate DARE on temporally ordered benchmarks with delayed feedback and compare it against common heuristic retraining strategies, including an action-rate-matched fixed-interval baseline designed to separate policy quality from update frequency. The goal of this evaluation is to test whether a sequential, operationally efficient retraining policy can maintain task performance while reducing unnecessary interventions under non-stationary data.

In summary, this paper makes three contributions:
(1) an operational formulation of retraining as a runtime-constrained sequential decision problem under delayed labels;
(2) a reproducible event-driven evaluation framework for policy comparison; and
(3) evidence that conservative contextual-bandit retraining outperforms practical heuristic retraining strategies, including an action-rate-matched fixed schedule, while reducing measured execution overhead (wall-clock runtime) under delayed feedback.

The remainder of the paper presents related work, system design, the experimental protocol, and results focused on Airline (a temporally ordered flight-delay classification stream), with Elec2 (an electricity market binary classification benchmark) and delayed-label stress tests for calibration and robustness.

\begin{figure*}[!t]
\centering
\includegraphics[
    width=\textwidth,
    trim=2mm 8mm 2mm 2mm,
    clip
]{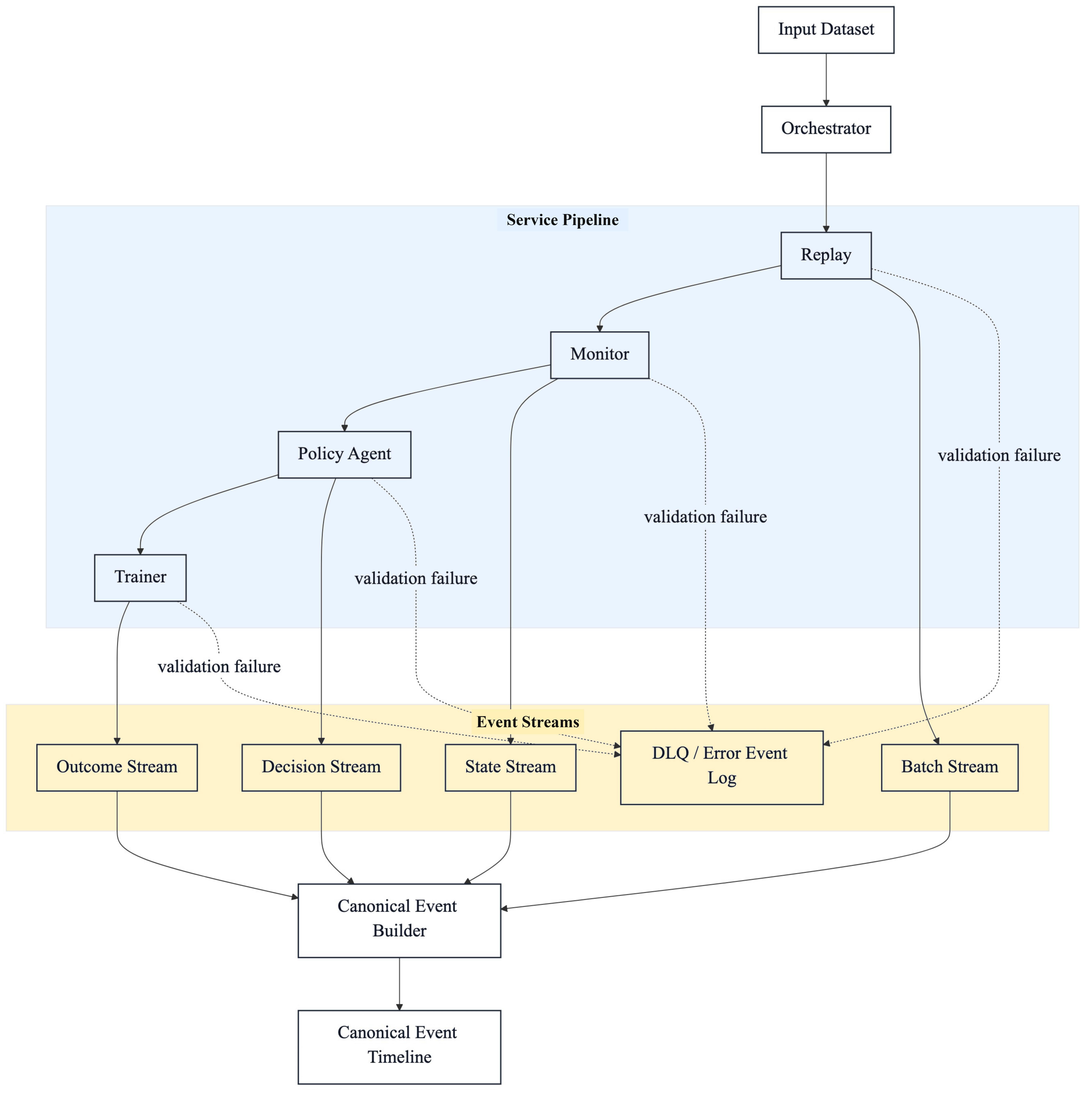}
\caption{DARE event-driven pipeline. Dashed arrows indicate validation-failure routing to the DLQ.}
\label{fig:dare_arch}
\end{figure*}

\section{Related Work}
\textbf{Drift detection and monitoring.}
The concept-drift literature has established robust taxonomies and detection mechanisms, ranging from classical surveys to recent systems work. Dong \emph{et al.} \cite{b7} and Böther \emph{et al.} \cite{b8} highlight systems-level considerations in drift-aware pipelines, including orchestration and data-centric controls.  Katalay \emph{et al.} \cite{b9} combine multiple monitoring criteria (e.g., distribution and divergence metrics) for automated retraining decisions.
However, these approaches primarily improve \emph{detection} or trigger design; they do not frame retraining as a sequential decision policy that explicitly trades off performance and intervention cost under delayed labels. Although contextual bandit methods provide a sequential decision mechanism \cite{b12}, they are typically studied for recommendation-style action selection rather than retraining-control decisions under drift, delayed labels, and explicit operational overhead.

\textbf{Retraining policy and operational cost.}
Chislett \emph{et al.} \cite{b2} propose AMUSE, an adaptive model-updating approach that learns retraining timing in a simulated drift environment while balancing predictive utility and operational overhead. Unlike fixed schedules or static trigger rules, AMUSE frames update timing as a learned sequential policy, though its evaluation is primarily simulation-based.
Regol \emph{et al.} \cite{b3} further analyze retraining timing and emphasize sensitivity to operating assumptions.
 Systems such as Modyn further highlight that the quality of retraining depends on data/pipeline choices and operational overhead, not just model metrics \cite{b8}. Still, most methods rely on fixed schedules, threshold rules, or static selection logic rather than learning a conservative action policy over time that is explicit about the action rate and budget. Action rate is the fraction of decision windows in which the policy triggers a model update.

\textbf{Sequential decision methods under delayed feedback.}
Rank \emph{et al.} \cite{b5} and Moothedath \cite{b6} motivate sequential decision-making under shifting environments and delayed feedback, but do not provide an end-to-end event-driven retraining orchestration framework with deterministic replay.
 In parallel, fairness-under-drift work shows that subgroup behavior can shift over time and should be monitored continuously \cite{b10}. Yet these lines are typically theoretical or domain-specific and are not integrated into an end-to-end retraining orchestration pipeline with deterministic replay, schema-level traceability, and direct comparison against practical threshold baselines.

\textbf{Positioning.}
Our work connects these strands by evaluating a conservative contextual-bandit retraining policy in an MLOps-oriented, event-driven framework, with explicit operationally efficient decisions and delayed-label robustness testing against standard heuristic retraining policies.

\section{System Design}
\label{sec:system}

\subsection{Design Objectives}
DARE is designed for policy-level retraining decisions under three practical constraints: non-stationary inputs, delayed labels, and bounded operational resources.  The core objective is to compare retraining policies under a shared execution protocol, not to optimize drift detection in isolation \cite{b3,b4,b7,b9}. Following this objective, the system enforces strict separation between (i) monitoring signals, (ii) policy decisions, and (iii) execution outcomes. Strict separation is enforced at the interface level: monitor emits state only, policy emits action only, and trainer emits outcome only. No component mutates another component’s artifacts in place; communication is event-based and append-only. This prevents signal leakage (e.g., future outcome information entering policy state), isolates failures to component boundaries, and ensures that policy comparisons reflect decision logic rather than hidden coupling between monitoring and execution stages.

\subsection{Event-Driven Runtime}
\label{subsec:runtime}
The implemented runtime is an offline, event-driven pipeline:
\begin{center}
\small
\begin{tabular}{@{}l@{}}
\texttt{replay $\rightarrow$ monitor $\rightarrow$ policy $\rightarrow$} \\
\texttt{trainer $\rightarrow$ canonical\_log}
\end{tabular}
\end{center}

Replay emits time-ordered batch and delayed-label events from CSV windows. Monitor consumes these events and emits state events. Policy consumes states and emits actions. Trainer consumes actions and emits cost/model-update outcomes. All stages write immutable JSON Lines (JSONL) events, which are merged into an append-only, canonical timeline for reproducible replay and post-hoc analysis \cite{b11}. This architecture follows the same systems motivation as recent MLOps pipeline work while making decision traces explicit \cite{b8,b9}.

\subsection{State, Action, and Outcome Contracts}
\label{subsec:contracts}

Each decision step is represented by a compact state record
\begin{equation}
\begin{aligned}
s_t = [&\text{drift features},\ \text{performance features},\\
      &\text{fairness features},\ \text{readiness flags},\ \text{metadata}]
\end{aligned}
\end{equation}

The implementation uses a schema-validated canonical state object (including IDs, window metadata, and model versioning), but policy decisions in this paper rely primarily on drift/performance readiness signals.

In the current implementation, \texttt{drift\_core} is the primary trigger signal; fairness fields are logged and available to policies, but are not required in the locked Airline operating point.  Here, ``locked'' means a fixed, pre-declared evaluation configuration (reward weights, action semantics, exploration settings, batching, and delay protocol) applied unchanged across all six seeds and baseline comparisons, with no per-run retuning.

Action semantics are intent-level:
\[
a_t \in \{\texttt{defer},\ \texttt{fine\_tune}\}
\]
for the LinUCB policy path. The execution layer also supports \texttt{full\_retrain} for heuristic control policies. For the locked Airline comparison, all updating policies execute the same \verb|fine_tune| operation. The \verb|full_retrain| action is implemented for extensibility and heuristic evaluation but is not used in the reported Airline operating point. Consequently, differences between policies reflect update timing rather than different update semantics. Outcomes are recorded through cost events (operational overhead, time, energy, pre-update metrics, fairness diagnostics) and optional model-update events when an update is executed.

\subsection{Policy Layer}
\label{subsec:policy}
We evaluate three deterministic baselines: no retrain, fixed interval, and threshold persistence, against one learned policy (LinUCB) \cite{b12}. 

The learned policy is conservative by design in three ways: (i) the action space is restricted to \{\texttt{defer}, \texttt{fine\_tune}\} (no aggressive full retrain action in the main RL path), (ii) reward shaping penalizes intervention cost and update frequency, and (iii) exploration is bounded by a fixed schedule to avoid unstable action-rate spikes. Given context vector $x_t$, LinUCB selects
\begin{equation}
a_t = \arg\max_{a \in \mathcal{A}}
\left(
\hat{\theta}_a^\top x_t + \alpha \sqrt{x_t^\top A_a^{-1}x_t}
\right),
\label{eq:linucb}
\end{equation}
where $\alpha$ controls exploration. We lock $\alpha=0.25$ and fix the exploration schedule across all runs to avoid run-specific retuning.

\subsection{Reward Construction with Delayed Feedback}
\label{subsec:reward}
Rewards are computed from logged outcomes, not from future information at decision time. Let $\Delta \tilde{p}_t$ denote normalized rolling performance change and $\tilde{c}_t$ normalized measured runtime. The base reward is
\begin{equation}
\begin{aligned}
r_t &= \Delta \tilde{p}_t - w_{\text{cost}}\tilde{c}_t
      - \lambda_{\text{act}}\mathbf{1}[a_t=\texttt{fine\_tune}] \\
    &\quad + \lambda_{\text{def}}\mathbf{1}[a_t=\texttt{defer} \land d_t<\tau_d].
\end{aligned}
\label{eq:reward}
\end{equation}

DARE is metric-agnostic at the orchestration level: the runtime, event contracts, and policy loop remain unchanged across tasks, while only the performance term $\Delta \tilde{p}_t$ is task-specific. In this paper, Airline uses rolling $\Delta$F1 (imbalanced classification), whereas Elec2 uses rolling $\Delta$Accuracy. This design allows the same decision framework to transfer across tasks with different primary performance metrics. An optional fairness penalty term is implemented but not enforced in the locked main result:
\begin{equation}
r_t^{\text{final}} = r_t^{\text{base}} - w_{\text{fair}}\tilde{g}_t
\label{eq:reward_fair}
\end{equation}
This keeps fairness measurable without over-claiming fairness-optimized control in the present phase \cite{b10}.

\subsection{Determinism and Reproducibility Controls}
\label{subsec:repro}
DARE enforces deterministic event IDs, schema validation, and dead-letter queue (DLQ) routing for invalid payloads. Canonical logs are append-only, and replay is file-order deterministic. Trainer execution is idempotent through action-level processed-index checks. These controls ensure that policy comparisons are attributable to policy logic and configuration, rather than orchestration artifacts.

\section{Experiments}
\label{sec:experiments}

\subsection{Experimental Questions}
We evaluate four questions:
(1) Can a conservative contextual-bandit retraining policy reduce operational overhead relative to standard heuristic triggers?
(2) Does this reduction preserve or improve task performance?
(3) How stable are outcomes across random seeds?
(4) Does policy behavior remain stable under delayed-label feedback?

\subsection{Dataset}
Our primary benchmark is Airline 120k, a temporally ordered binary classification stream for predicting arrival delay $>15$ minutes \cite{b13}. The stream is class-imbalanced (positive class frequency $\approx 17.5\%$), so we report F1 alongside balanced accuracy. Elec2 is a temporally ordered binary stream benchmark derived from electricity market signals \cite{b14,b15}. In our calibration runs, Elec2 is used as secondary calibration evidence rather than as a headline-claim-setting benchmark. Data are processed in fixed batches of 200 for all compared policies.

\subsection{Setup and Protocol}
We compare three deterministic baselines (\textit{no retrain}, \textit{fixed interval}, and \textit{threshold persistence}) against a locked LinUCB policy with action space $\{\texttt{defer}, \texttt{fine\_tune}\}$ \cite{b3,b4}. In the reported Airline experiments, all updating policies execute the same \verb|fine_tune| operation. Consequently, observed differences arise from update timing and frequency rather than different retraining mechanisms. The no-retrain baseline never updates the model after initialization. These baselines are intended to represent simple operational policies commonly used when learned retraining control is unavailable.

To distinguish adaptive decision quality from simply updating less frequently, we additionally evaluate an action-rate-matched fixed-interval baseline whose interval is selected solely to match the locked LinUCB intervention frequency before evaluation. All reward parameters, exploration settings, and baseline hyperparameters are fixed prior to the six-seed evaluation. 

For Airline, threshold persistence uses drift threshold $T=1.8$ and persistence $K=3$: retraining is triggered only when \texttt{drift\_core} exceeds $T$ for $K$ consecutive decision windows; otherwise the policy defers. The threshold-persistence hyperparameters ($T=1.8$, $K=3$) were selected through our own predefined baseline calibration on development runs (separate from the final six-seed comparison table). We evaluated candidate settings by their stability--coverage behavior (avoiding degenerate always-update/never-update patterns) and then froze the selected pair before final policy comparison to avoid outcome chasing.

The underlying Airline predictor is an incremental linear classifier implemented with \texttt{SGDClassifier} using logistic loss. Before each eligible update, the current model is evaluated on the incoming labeled window; the reported performance metrics therefore describe pre-update behavior and avoid evaluating the model on a window after training on it. An executed \texttt{fine\_tune} applies one \texttt{partial\_fit} update to that window using inverse-frequency class weights and then stores the resulting model as a new version. If delayed labels are unavailable, the requested update is not executed. This design isolates policy quality from differences in model architecture or optimization procedure. All policies therefore share identical model architecture, feature representation, windowing protocol, and delayed-label configuration; only the retraining decision policy differs. All compared policies use this identical
predictor, feature pipeline, and update procedure, so their only experimental difference is when they request an update.

\subsection{Metrics}
We report cumulative update runtime, F1, balanced accuracy, action rate, and update count. Runtime captures the operational overhead directly controlled by the retraining policy, measured via wall-clock execution time rather than fixed proxy-cost assumptions, while F1 and balanced accuracy measure predictive performance under class imbalance. Action rate and update count quantify intervention aggressiveness. While balanced accuracy remains close to chance for all policies, indicating the underlying classifier has limited overall discriminative power on this task, the relative precision/recall differences across retraining policies (Table I) still meaningfully distinguish policy quality, since all policies share the identical underlying model and differ only in retraining timing.                                                                    

\textbf{Runtime  accounting.}
We measure cumulative wall-clock update runtime for each retraining action executed under a fixed hardware and software configuration. Because all compared policies execute the same update routine, cumulative runtime provides a directly observed measure of operational overhead. Under usage-based cloud pricing, infrastructure costs scale approximately with compute time for a fixed deployment configuration; runtime therefore serves as the primary experimental resource metric while avoiding provider-specific billing assumptions. 

Monitoring, drift computation, and state construction are executed for every evaluated policy. Because these operations are identical across compared policies, their computational overhead is treated as constant and excluded from comparative runtime analysis. 

\begin{table*}
\caption{Airline 120k policy comparison (6 seeds, mean $\pm$ SD)}
\label{tab:airline_main_final}
\centering
\small
\setlength{\tabcolsep}{2.8pt}
\begin{tabular}{@{}lllllll@{}}
\toprule
Policy                       & Runtime(s)        & F1              & Precision       & Recall          & Action rate    & Balanced accuracy    \\ \midrule
No retrain                   & 0.0000$\pm$0.0000 & 0.0000$\pm$0.0000 & 0.0000$\pm$0.0000 & 0.0000$\pm$0.0000 & 0.0000$\pm$0.0000 & 0.0000$\pm$0.0000 \\
Fixed (N=10)                 & 2.2936$\pm$1.3266 &0.1724$\pm$0.0114 &0.1309$\pm$0.0076 & 0.4800$\pm$0.0431 & 0.0991$\pm$0.0000 & 	0.5089$\pm$0.0033\\
Fixed matched (N=64)         & 0.3870$\pm$0.2118 & 0.1742$\pm$0.0278 & 0.1258$\pm$0.0182 & 0.4990$\pm$0.0901 & 0.0154$\pm$0.0000 & 0.5026$\pm$0.0053\\ 
Threshold persistence (1.8,3) & 1.8466$\pm$0.2802 & 0.1547$\pm$0.0543 & 0.1167$\pm$0.0290 & 0.4465$\pm$0.1993 & 0.1145$\pm$0.0000 & 0.5002$\pm$0.0041\\
LinUCB (locked)              & 0.2687$\pm$0.0859 & 0.2065$\pm$0.0254 & 0.1596$\pm$0.0179 & 0.6012$\pm$0.1310 &  0.0157$\pm$0.0036 & 0.5057$\pm$0.0059 \\ \bottomrule
\end{tabular}
\end{table*}

\begin{figure}[t]
\centering
\includegraphics[width=\linewidth]{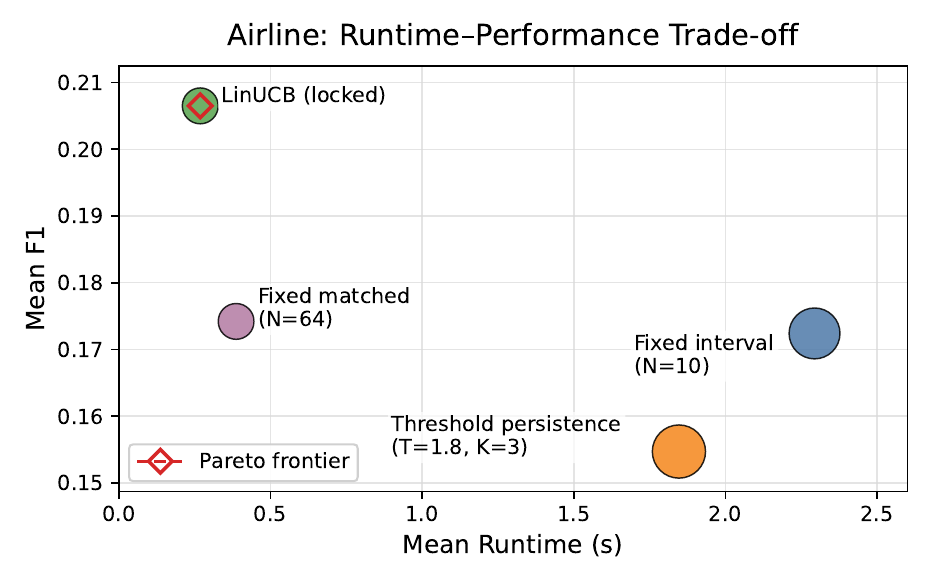}
\caption{Airline Pareto trade-off (cumulative update runtime vs. F1). Each bubble represents a retraining policy, with bubble area proportional to action rate; larger bubbles indicate more frequent update decisions. The red diamond marks the non-dominated Pareto-front policy. LinUCB occupies the favorable low-runtime/high-F1 region.}
\label{fig:airline_pareto}
\end{figure}

\begin{table*}[!t]
\centering

\begin{minipage}[t]{0.48\textwidth}
\centering
\caption{Delayed-label stress test (Airline, LinUCB locked; means over 6 seeds)}
\label{tab:delay_stress_airline}
\small
\setlength{\tabcolsep}{3pt}
\begin{tabular}{@{}cccccc@{}}
\hline
\textbf{Delay} & \textbf{Act. rate}  & \textbf{Updates} & \textbf{F1} & \textbf{Reward} \\
\hline
0  & 0.0157  & 9.17 & 0.2058 & -0.0010 \\
2  & 0.0157  & 9.17 & 0.2065 & -0.0010 \\
5  & 0.0157  & 9.17 & 0.2076 & -0.0010 \\
10 & 0.0157  & 8.17 & 0.2229 & 0.0015 \\
\hline
\end{tabular}
\end{minipage}
\hfill
\begin{minipage}[t]{0.48\textwidth}
\centering
\caption{Metric alignment ablation (Airline, LinUCB; means over 6 seeds)}
\label{tab:metric_ablation}
\small
\setlength{\tabcolsep}{3pt}
\begin{tabular}{@{}lcccc@{}}
\hline
\textbf{Reward metric} & \textbf{Action rate} & \textbf{Updates} & \textbf{F1} \\
\hline
$\Delta$Accuracy & 0.2886 & 168.83 & 0.1813 \\
$\Delta$F1       & 0.0143 & 8.33  & 0.2290 \\
\hline
\end{tabular}
\end{minipage}

\end{table*}

Across six seeds, the locked LinUCB policy achieves the highest mean F1 while maintaining the lowest operational overhead among adaptive policies. Compared with fixed-interval retraining, cumulative update runtime is reduced by 88.3\%, while maintaining higher F1. Relative to threshold persistence, runtime decreases by 85.4\%. Even when compared with an action-rate-matched fixed schedule, LinUCB requires 30.6\% less cumulative update runtime while preserving superior predictive performance, indicating that the observed gains cannot be explained solely by updating less frequently. 

Under label delays of 0--10 batches (Table~\ref{tab:delay_stress_airline}), action rate remains flat at 0.0157 with no collapse to always-update or never-update behavior. At delay 10, updates decrease (9.17 to 8.17) and F1 increases (0.2058 to 0.2229), consistent with a more conservative intervention pattern in an imbalanced setting.

Unless otherwise noted, headline values refer to the locked operating-point evaluation (Table~\ref{tab:airline_main_final}); metric-alignment ablation values come from dedicated 6-seed reruns and are interpreted as within-ablation contrasts.

\section{Results and Discussion}
\label{sec:results}

\textbf{Airline as the primary result.}
Table I shows that LinUCB achieves the highest F1 among adaptive policies. The action-rate-matched baseline demonstrates that this improvement is not explained solely by updating less often, while runtime measurements confirm the same ordering under a directly observed operational metric. 

\textbf{Metric alignment drives behavior.}
Table~\ref{tab:metric_ablation} shows that the choice of reward metric substantially changes policy behavior under class imbalance. Optimizing 
$\Delta$Accuracy causes LinUCB to over-intervene (168.83 updates) and yields lower F1 (0.1813), since majority-class accuracy dominates the reward signal. Optimizing $\Delta$F1 instead restores conservative behavior (8.33 updates) and improves minority-class performance (F1 0.2290). This contrast confirms that DARE's orchestration is metric-agnostic, but the resulting policy behavior is shaped entirely by the chosen reward objective. (Absolute values in this ablation come from dedicated 6-seed reruns; the comparison is a within-table contrast between reward definitions, not against the main Table~\ref{tab:airline_main_final} results.)

A complementary reward-mode ablation (raw vs. rolling) shows that rolling smoothing reduces reward variance (0.1698 to 0.0952) while leaving fine-tune rate unchanged (0.0600), indicating a more stable learning signal without increased intervention.

\textbf{Delayed-label robustness.}
Table~\ref{tab:delay_stress_airline} shows stable behavior across delays of 0--10 batches: action rate is unchanged at 0.0157, cost changes modestly (13.75 to 12.25), and there is no collapse to degenerate always-update or never-update regimes. The F1 increase at delay 10 coincides with fewer updates (9.17 to 8.17), consistent with reduced over-intervention under delayed feedback in an imbalanced setting. Reward remains near zero across delays (from -0.0010 at delays 0--5 to 0.0015 at delay 10), indicating no instability or collapse in the delayed-feedback regime.

\textbf{Elec2 as calibration evidence.}
Elec2 exhibits the same directional behavior as Airline: heuristic policies intervene substantially more frequently than LinUCB, supporting the consistency of the learned policy across temporally ordered streams. Because Elec2 is used as calibration rather than the headline benchmark, we report it as supporting evidence rather than as the primary policy comparison.

\textbf{Limitations.}
Several limitations bound the current results. Runtime measurements were collected under a single hardware configuration. Although cumulative runtime provides a directly observed measure of operational overhead, absolute execution time may vary across deployment environments. The offline replay protocol does not capture queue lag, distributed failure modes, or live serving effects. In addition, the locked LinUCB action space in the main result is binary (defer vs.\ fine-tune), while broader action sets under explicit budget constraints remain future work. The evaluation uses two temporally ordered tabular datasets; generalization to image streams, text streams, or high-dimensional feature spaces with different drift characteristics remains an open question. Fairness and energy signals are logged and reported for diagnostics, but they are not enforced as primary optimization objectives in the main Airline result. Finally, the six-seed evaluation, while stronger than single-run reporting, does not yet provide formal regret bounds or convergence guarantees for the conservative operating regime.

\section{Conclusion}
\label{sec:conclusion}
This paper reframes model retraining under drift as a sequential decision problem with delayed feedback, rather than a static trigger-design problem. We presented DARE, an event-driven MLOps architecture with schema-validated events and deterministic replay that enables reproducible policy-level comparison under shared execution conditions.

Empirically, on Airline 120k, a locked conservative LinUCB policy achieves a favorable runtime--performance operating point versus heuristic retraining baselines while remaining stable under delayed labels. These results support the central design claim: policy quality depends on explicit action-cost trade-offs under delay, not drift detection alone. They also expose two methodological lessons that generalize beyond the Airline benchmark.

First, threshold-based retraining policies are not configuration-invariant across datasets: a threshold setting calibrated for Airline did not transfer reliably to Elec2 without re-tuning, indicating that heuristic trigger policies require dataset-specific recalibration for stable operation. Second, DARE is reward-metric agnostic at the orchestration level, but policy behavior is strongly shaped by the chosen reward definition, as shown in Section IV, optimizing $\Delta$Accuracy led to over-intervention. These findings reinforce that deployment-ready retraining policies need both calibrated trigger logic and metric-task alignment.

\textbf{Future work.} Next steps are to couple fairness signals directly to action selection, expand evaluation across more datasets and seeds, and extend from offline replay to online validation with richer budget-aware action spaces.

\vspace{12pt}

\end{document}